# An Image dehazing approach based on the airlight field estimation

Lijun Zhang          Yongbin Gao          Yujin Zhang

Shanghai, China

## Abstract

*This paper proposes a scheme for single image haze removal based on the airlight field (ALF) estimation. Conventional image dehazing methods which are based on a physical model generally take the global atmospheric light as a constant. However, the constant-airlight assumption may be unsuitable for images with large sky regions, which causes unacceptable brightness imbalance and color distortion in recovery images. This paper models the atmospheric light as a field function, and presents a maximum a-priori (MAP) method for jointly estimating the airlight field, the transmission rate and the haze free image. We also introduce a valid haze-level prior for effective estimate of transmission. Evaluation on real world images shows that the proposed approach outperforms existing methods in single image dehazing, especially when the large sky region is included.*

1. Introduction

Imaging outdoor is often degraded by atmospheric scattering caused by haze, fog and mist in bad weather. This is due to the absorption of aerosol particles and the airlight scattering [1]. Poor visibility of hazy images is also a major problem for many computer vision applications such as surveillance, intelligent vehicles and outdoor object recognition. It makes image dehazing to be a very important and challenging task.

Image dehazing algorithms can be divided into two main types. First is based on the atmospheric scattering model. Second is the non-model based algorithm which includes histogram equalization, homomorphic filtering and retinex [2-4]. Model-based algorithms are more reliable as they exploit the underlying physics of the degradation process, and there has been considerable research in it. In the very beginning, the method depended on knowledge of aerosol pollution level or scene geometry [5,6]. The method in [7] requires some depth information from user inputs or known 3D models. On purpose of evaluating the scene depth, many methods have been proposed by using two or more images in the same scene yet at different times or different degrees of polarization [8-10].

Narasimhan[11] first addresses the question of how to dehaze a single image without using precise weather or depth information. Tan[12] presents the first automatic single image dehazing method. His main idea is to estimate and remove the airlight component and then enhance the local contrast of the restored image, without estimation of scene depth. Fattal[13] proposes an approach to estimate the scene transmission and intensity under the assumption that they are locally uncorrelated. He[14] proposes the prior-dark channel prior to remove haze from a single input image.

The single image dehazing algorithms based on energy minimization attracted much attention since [15], which introduced a Bayesian probabilistic method that jointly estimates the scene albedo and depth with energy minimization of a factorial Markov Random Field (fMRF). In [16], Meng proposes a context regularization method together with the transmission boundary constraint. Wang[17] proposes a Bayesian approach for the adaptive fusion of depth maps from multiple scales. Zhou[18] proposes a constrained energy formulation under Bayesian and variation theories.

In the previous works, little attention has been paid to the estimation of the global atmospheric light *A*. Many of them take the color of the most haze-opaque region as the *A*'s value [12,14,16,17], or as an initial guess of an energy minimizing procedure [13,19]. In [20], Sulami estimates *A* in two steps: at the first step its orientation is estimated under a geometric constraint; at the second step, its amplitude is estimated under a global image prior. A similar method for estimating *A* is used in [18]. All previous works determine the atmospheric light under the assumption that it is a constant. However, this assumption may be unsuitable when there is an uneven light source, e.g., when the image covers large sky and the sunlight is very influential. To address this issue, we take the atmospheric light as a 2-dimensional field function, which is called as the airlight field *A*(x), and present a method to estimate it.

Additionally, many previous methods take the estimated transmission to recover the haze free image by directly dividing it from the observed image [14,16,17,20]. While all these methods achieve remarkable results, they suffer from halo and staircase artifacts due to noise and quantization error. This phenomenon frequently occurs for the image with a large sky region where the transmission is close to zero.

In this paper, we formulate the single image dehazing as a joint estimation of the triple variables, i.e., the airlight field, the transmission rate and the true image, with energy minimization of a Markov Random Field (MRF). The airlight field is represented by a linear combination of a set of smooth basis functions, which ensures the smoothly varying property of $A(x)$. Moreover, we found that the proposed approach can alleviate the halo and staircase effects in the restored image which are caused by numerical issues. To estimate the transmission accurately, we further derive an analytic expression with regard to the dependence between the pixel transmission and its fogging degree. This is called as haze-level prior and is modeled into an optimization problem to determine the unknown transmission. In comparison with several relevant approaches, the proposed method outperforms them in single image dehazing, especially for images with large sky regions.

The remainder of this paper is organized as follows. In Section 2, we first review some well known single image dehazing algorithms and their limitations. In Section 3, we propose the airlight field model and the energy minimization framework with the haze-level prior. In Section 4, we describe the joint MAP estimation for the airlight field, the transmission and the true image. The implementation and results of our approach are given in Section 5, followed by some discussions in Section 6. It is summarized in Section 7.

2. Related works

In [8], Nayar derives a dichromatic atmospheric scattering model that formulates the observed hazy Image $I$ to be a composite of the attenuated component $L_s$ and the airlight component $L_a$,

$$I(x) = L_s + L_a$$
$$= \frac{L_\infty \rho(x) e^{-\beta d(x)}}{d^2(x)} + L_\infty (1 - e^{-\beta d(x)}) \quad (1)$$

where $d(x)$ and $\rho(x)$ are the scene depth and albedo respectively, at the scene point corresponding to image coordinates x. As a vector in RGB space, $L_\infty$ is the airlight for image points corresponding to scene points at infinity, i.e., the global atmospheric light. The attenuation coefficient $\beta$ does not depend on wavelength for particles of fog and haze, and is assumed to be uniform across the entire scene.

Let $J(x) = \frac{L_\infty \rho(x)}{d^2(x)}$ represent the wanted haze free image, $t(x) = e^{-\beta d(x)}$ represent the transmission, and denote $L_\infty$

as $A$, then Equation (1) can be rewriten as

$$I(x) = t(x)J(x) + (1 - t(x))A \quad (2)$$

In [14], He proposes the dark channel prior, and give the estimate of transmission as

$$\hat{t}(x) = 1 - \omega \min_{y \in \Omega_x} \left( \min_{c \in \{r,g,b\}} \frac{I^c(y)}{A^c} \right) \quad (3)$$

This is taken as the initial estimates for the matting algorithm that refines the transmission map. The final true image is recovered by

$$J(x) = \frac{I(x) - A}{\max(t(x), t_0)} + A \quad (4)$$

where $t_0$ is a tiny constant in case of $t(x)$ is very close to zero.

In [15], Kratz formulate an energy function framework for joint optimization of the scaled albedo $\tilde{\rho}_x = \frac{\rho(x)}{d^2(x)}$ and scaled depth $\tilde{d}_x = \beta d(x)$. They build up complete MRFs that can be optimized using the existing inference approaches, producing the estimated values of $\tilde{\rho}$ and $\tilde{d}$. The final dehazing result is just ordinary, but the energy minimizing model is generic and can easily be refined to introduce other constraints which lead to better results.

In [16], Meng first get the initial estimate of the transmission by exploiting an inherent boundary constraint,

$$t_b(x) = \min \left\{ \max_{c \in \{r,g,b\}} \left( \frac{A^c - I^c(x)}{A^c - C_0^c}, \frac{A^c - I^c(x)}{A^c - C_1^c} \right), 1 \right\} \quad (5)$$

Then they find an optimal transmission by minimizing an energy function.

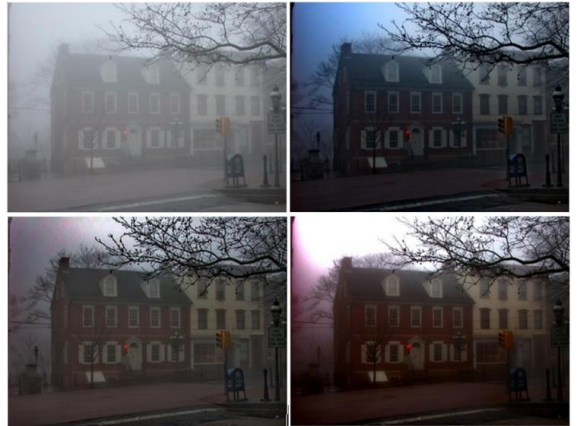

Figure 1: Brightness imbalance and color distortion of the recovery image due to non-constant atmospheric light . TopLeft: input hazy image. TopRight – BottomLeft – BottomRight: recovery images by [14]-[16]-[20].

Above methods either rely on user input or use most haze-opaque region to estimate the global atmospheric light. In [20], Sulami present an automatic method to recover $A$, which is consequently used to dehaze the single image via the dark channel prior. Nevertheless, all previous works recover *A* under the assumption that it is constant throughout the image. However, this assumption may be unsuitable when the image covers large sky and the sunlight is very influential. As said in [14], when the sunlight is strong, if we regard *A* as a constant, the recovery image will have uneven brightness. Moreover, the false estimate of *A* causes that recovery images suffer from brightness imbalance and color distortion, as shown in Figure 1.

Additionally, many previous methods remove haze in two stages, that are estimate of transmission and subsequently solving Equation (4) to recover the haze free image [14,16,17,20]. In comparison with the joint MAP recovery of the image, this scheme is prone to noise and quantization error when $t(x)$ is closed to zero. When the image contains large sky region, it will induce terrible halo and staircase artifacts, as shown in Figure 2.

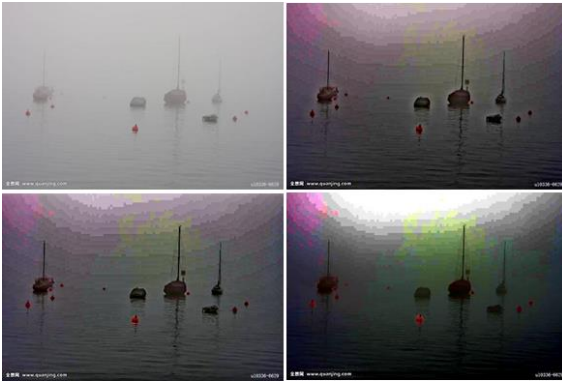

Figure 2: Halo and staircase effects in large sky region. TopLeft: input hazy image. TopRight – BottomLeft - BottomRight: recovery images by [14]-[16]-[20].

In brief, the above methods cannot handle hazy images which contain large sky region and may fail when the constant-airlight assumption is broken. Unfortunately, most outdoor photos contain sky regions that cannot be ignored.

3. Problem formulation

In this section, we first propose the airlight field model. Then an energy minimization framework is presented. Finally we introduce the haze-level prior for transmission estmate.

3.1. Airlight Field (ALF)

Many previous works use the color of the most haze-opaque region in the sky as the global atmospheric light. However, as pointed out in [8], we should only have the value of *A* at the horizon at an infinite distance. This implies that haze is homogeneous only in the horizontal direction.

As shown in Figure 3, haze aerosol and other suspended particles in the atmosphere are always distributed near the surface of the earth. The vertical thickness of the haze is about hundreds of meters, which is assumed to be *H*. When the elevation angle is relatively large, the thickness of the observed haze is limited. Under the assumption that there is still a constant *A* or $L_\infty$, the airlight component at point *P* should be

$$L_a = L_\infty\big(1 - e^{-\beta d(P)}\big) + L_{solar}(P) \qquad (6)$$

where $L_{solar}$ is the brightness caused by direct sunlight. It is not uniform. Generally, the location near the sun has greater intensity, e.g., $L_{solar}(P) < L_{solar}(Q)$. Because the distance from point *P* to the observer $d(P) < \infty$, we have $L_a \neq L_\infty$. If the influence of $L_{solar}$ is ignored, then $L_a < L_\infty$. On the other hand, if the sunlight is very influential, there may be $L_a > L_\infty$. That is to say, due to the direct impact of the sun, usually the sky area has greater brightness, so it is more likely to be used as the estimate of *A* by many of the above algorithms. The result is that when the image contains large sky regions, these fog removal algorithms may fail due to the false estimate of *A*, as shown in Figure 1.

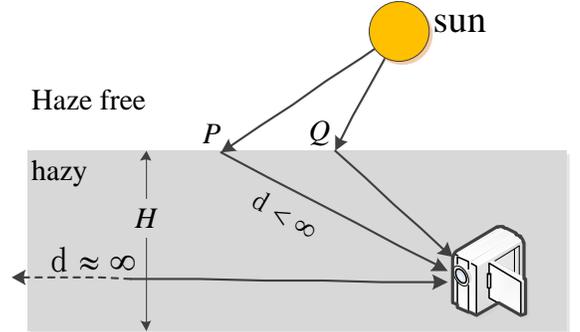

Figure 3: Non-constant airlight due to haze inhomogeneity and influential sunlight.

In order to compensate for the influence of non-uniform haze and direct sunlight, we define the global atmospheric light as a two-dimensional field function *A*(x). Thus Equation (2) is modified as

$$I(x) = t(x)J(x) + (1 - t(x))A(x) \qquad (7)$$

where $A(x)$ is called as airlight field, which is smooth and slowly varying.

3.2. Energy minimization

In order to restore the true image *J*(x), we use energy minimization to jointly estimate *t*, *J* and *A*. The simplest form of the energy minimization problem is

$$\text{minimize } \|\nabla t\|_p, \ \|\nabla J\|_p, \ \|\nabla A\|_p \qquad (8)$$
$$s.t. \ \left\| I(x) - \big(t(x)J(x) + (1 - t(x))A(x)\big) \right\|_2^2 = 0$$

where $\|\cdot\|_p$ is the *p*-norm, $\nabla(\cdot)$ is the gradient operator. This is a multi-objective constrained optimization problem. If we let

$$\|\nabla(\cdot)\|_p = \int_\Omega \rho(\cdot)dx \tag{9}$$

then the Lagrange dual of (8) is

$$\text{minimize } E = \int_\Omega \left(I(x) - t(x)J(x) - (1-t(x))A(x)\right)^2 dx$$
$$+ \lambda_1 \int_\Omega \rho(t)dx + \lambda_2 \int_\Omega \rho(J)dx + \lambda_3 \int_\Omega \rho(A)dx \tag{10}$$

where $\lambda_1$, $\lambda_2$ and $\lambda_3$ are regularization factors, $\Omega$ represents the whole image domain. It means that the data term energy is minimized under the assumption that $t(x)$, $J(x)$ and $A(x)$ satisfy the smoothness constraint. If the global atmospheric light is assumed to be a constant, then the regularization term of $A(x)$ can be ignored, which is essentially similar to [15].

As for the regularization terms $\|\nabla t\|_p$, $\|\nabla J\|_p$ and $\|\nabla A\|_p$, we consider three types of norm operator.

A) $L^2$-norm

When $\rho(u) = |\nabla u|^2 = u'^2_x + u'^2_y$, the regularization terms are $L^2$-norm and the algorithm becomes the least square (LS) estimation where closed form solutions are easily computed. Though the quadratic optimization problem can be solved very efficiently, the usefulness of $L^2$-norm regularization is very limited. The results are likely to be contaminated by Gibbs's phenomenon and smoothing of edges.

B) TV-norm

When $\rho(u) = |\nabla u| = \sqrt{u'^2_x + u'^2_y}$, the regularization terms become total varation(TV) norm that belongs to $L^1$-norm. TV regularization with non-constant regularization parameter has already been studied in several articles [21-23]. The main advantage is that their solutions can avoid ringing and preserve edges very well, but there are computational difficulties.

C) SAD-norm

For simplicity, we use another type of $L^1$-norm by setting $\rho(u) = |u'_x| + |u'_y|$ in this paper, which can also be expressed as the sum of absolute differences(SAD)

$$\rho(u(x)) = \sum_{y \in N_x} |u(x) - u(y)| \tag{11}$$

where $N_x$ represent the neighborhood of the pixel x.

### 3.3. Haze-level Prior

It is difficult to get the accurate estimate of $t(x)$ by directly optimizing Equation (10), unless other effective regularization terms are introduced. In this section, we propose a haze-level prior based on the relationship between transmission rate and fogging degree, which greatly improves the accuracy of transmission.

While attenuation causes scene intensity to decrease with pathlength, the airlight component $L_a$ increases with pathlength. It therefore causes the apparent brightness of a scene point to increase with depth. As shown in Figure 4, in the region with heavy haze which is marked with red box, the transmission rate should be relatively low, and vice versa. This constraint reflects the relationship between transmission rate and fogging degree. Thus, we can define a cost function $\mathcal{F}(x)$, which should have such a property: for the heavy hazy region in the scene, $\mathcal{F}(x)$ is incremental with $t(x)$; while for the light hazy region, $\mathcal{F}(x)$ is declining. The fogging degree at a point x can be expressed by the proximity between the intensity $I(x)$ and the airlight $A(x)$. Hence, we construct such a cost function

$$\mathcal{F}(x) = \sum_c |R^c(x)|^2 \tag{12}$$

$$R^c(x) = t(x) - 1 + \frac{I^c(x)}{A^c(x)} \tag{13}$$

where $c \in \{r, g, b\}$ is color channel index, and transmission rate $0 \leq t(x) \leq 1$.

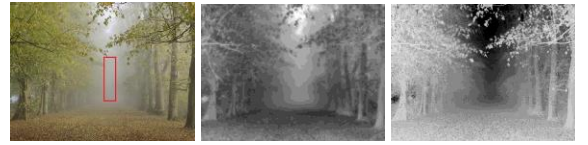

Figure 4: Relationship between transmission and fogging degree. Left: Input hazy image. The most hazy region is marked with red box. Middle: estimated transmission map without haze-level prior. Right: estimated transmission map with haze-level prior.

Usually the airlight field $A(x)$ has a higher intensity. If the intensity at point x is closer to the airlight, i.e. $I(x) \approx A(x)$, then it is likely to be heavily hazy, and $\mathcal{F}(x)$ is incremental with $t(x)$; if $I(x) \ll A(x)$, $\mathcal{F}(x)$ is declining with $t(x)$. As shown in Figure 5, the above defined $\mathcal{F}(x)$ can meet the requirement of the prior.

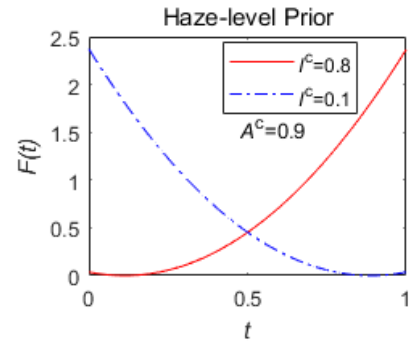

Figure 5: Cost function on haze-level prior. $I^c = 0.8$ represents a pixel with heavy haze and $I^c = 0.1$ represents a pixel with light haze.

Furthermore, we found through experiments that a better defogging effect can often be obtained by modifying Equation (13) into the following form

$$R^c(x) = t(x) - 1 + \frac{\min_c I^c(x)}{A^c(x)} \quad (14)$$

If we let the derivative of $\mathcal{F}(x)$ to $t(x)$ to be zero, then this quadratic function is minimized and we get the optimal transmission

$$\hat{t}(x) = 1 - \operatorname*{mean}_c \left( \frac{\min_c I^c(x)}{A^c(x)} \right) \quad (15)$$

It is similar in form to Equation (3) which refers to a estimate of the transmission as the dark channel prior [14]. This may illustrate the rationality of Equation (14). However, the dark channel prior is computed in local regions resulting in blocky estimates, while the haze-level prior is pixel-wise resulting in finer estimates.

## 4. Airlight field estimation

In this section, we first derive the energy minimization formula for the case where the global atmospheric light is a constant, i.e., the scene has a constant background radiation (CBR). Then we present the method for the airlight field (ALF) estimation where the atmospheric light is no longer a constant. Finally, we describe the whole procedure of the proposed method.

### 4.1. Constant background radiance

To facilitate the description, we first consider the case where the global atmospheric light is a constant, i.e., $A(x)=A_0$. While many previous works take the color of the most haze-opaque region as the value of $A_0$, we introduce a MAP method for estimating $A_0$ in this section. First of all, we add into Equation (10) a regularization term $E_{HL}$ which represents the haze-level prior,

$$\operatorname*{minimize}_{t,J,A_0} E = E_d + \lambda_1 E_t + \lambda_2 E_J + \lambda_3 E_{HL} \quad (16)$$

$$= \sum_{x,c} \left( I^c(x) - t(x) J^c(x) - (1-t(x)) A_0^c \right)^2 +$$

$$\lambda_1 \sum_x \rho(t(x)) + \lambda_2 \sum_{x,c} \rho(J^c(x)) + \lambda_3 \sum_{x,c} |R^c(x)|^2$$

where $E_d$ is the data term, $E_t$, $E_J$ and $E_{HL}$ are three regularization terms, $\rho(\cdot)$ is defined in Equation (11), and $R^c(x)$ is defined in Equation (13) or (14). We can solve (16) by alternatively minimizing $t(x)$, $J(x)$ and $A_0$ while fixing the other variables at each iteration. Note that given $t(x)$ and $J(x)$, the subproblem of solving $A_0$ can be formulated by the least square method for $c \in \{r,g,b\}$,

$$A_0^c = \frac{\sum_x (I^c(x) - t(x) J^c(x))(1-t(x))}{\sum_x (1-t(x))^2} \quad (17)$$

The other two variables $t(x)$ and $J^c(x)$ are recovered alternately by the gradient descent method. Considering that regularization of $E_J$ can smooth the true image and weaken the edge, we give it a small regularization factor. The following equations provide closed form expressions for the gradients of the energy function $E$ w.r.t. parameters $t(x)$ and $J^c(x)$ for SAD-norm:

$$\frac{\partial E}{\partial t(x)} = 2 \sum_c \left( I^c(x) - t(x) J^c(x) - (1-t(x)) A_0^c \right) \left( A_0^c - J^c(x) \right)$$

$$+ 2\lambda_1 \sum_{y \in N_x} \operatorname{sgn}(t(x) - t(y)) + 2\lambda_3 \sum_c R^c(x) \quad (18)$$

$$\frac{\partial E}{\partial J^c(x)} = 2 \left( I^c(x) - t(x) J^c(x) - (1-t(x)) A_0^c \right) (-t(x))$$

$$+ 2\lambda_2 \sum_{y \in N_x} \operatorname{sgn}(J^c(x) - J^c(y)) \quad (19)$$

where $\operatorname{sgn}(\cdot)$ is the sign function. The recovered transmission map and the output image are shown in Figure 6. The recovery image $J$ still has brightness imbalance due to the large sky region included in the image. Nevertheless, our result has a good color fidelity compared with other algirithms whoses results are shown in Figure 1.

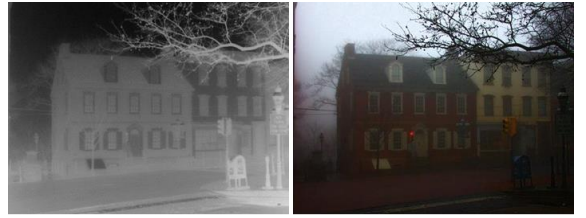

Figure 6: CBR Estimation. Left: transmission map. Right: recovery image.

### 4.2. Basis function fitting

The airlight field $A(x)$ is inhomogeneous under the following circumstances: 1) the image contains large sky area; 2) the scene is affected by strong sunlight; 3) the haze particles are not uniform; 4) there are additional local light sources in the scene. Inspired by [24], we assume $A(x)$ to be smoothly varying, and represent it as a linear combination of a given set of basis functions. Thus, $A(x)$ can be expressed in the following form

$$A^c(x) = \sum_{i=0}^{M-1} w_i^c g_i(x) \quad (20)$$

where $c \in \{r,g,b\}$, $w_0 \cdots w_{M-1}$ are weighting coefficients, $g_0 \cdots g_{M-1}$ are 2-dimensional smooth basis functions which ensure the smoothly varying property of the airlight field. As in [24], the airlight field theoretically can be approximated by a linear combination of a number of basis functions up to an arbitrary accuracy, given a sufficiently large number $M$ of basis functions. The estimation of the airlight field is transformed into finding the optimal coefficients $w_0 \cdots w_{M-1}$ in Equation (20).

In the implementation of the proposed method, 5 orthogonal Legendre polynomial functions, which are 2-order precision, are used to approximately estimate the airlight field, i.e., $M = 5$ in this paper. For an $n$-order Legendre polynomial function, the number of basis functions is related to the order $n$, i.e., $n = 2,3,4$ corresponds to $M = 5,10,15$ respectively. If $M = 1$, the airlight field function $A(x)$ is degenerated to a constant $A_0$.

In Equation (20), the subscript $i = 0$ represents the DC component. Considering that the airlight field is slowly varied, we must regularize its non-DC components to prevent a dramatic fluctuation. Thus, an additional $\|w_{i\neq 0}\|_2^2$ term is introduced into the energy function,

$$\begin{aligned} \underset{t,J,w}{\text{minimize}} E &= E_d + \lambda_1 E_t + \lambda_2 E_J + \lambda_3 E_{HL} + \lambda_4 \|w_{i\neq 0}\|_2^2 \\ &= \sum_x \sum_c \left(I^c(x) - t(x)J^c(x) - (1-t(x))A^c(x)\right)^2 \\ &\quad + \lambda_1 \sum_x \rho(t(x)) + \lambda_2 \sum_{x,c} \rho(J^c(x)) \\ &\quad + \lambda_3 \sum_{x,c} |R^c(x)|^2 + \lambda_4 \sum_{i=1}^{M-1} |w_i^c|^2 \end{aligned} \quad (21)$$

where the first four terms is the same as in Equation (16), except that the constant $A_0^c$ is changed to $A^c(x)$. The gradient descent method is used to find the optimal $t(x)$ and $J(x)$. The partial derivatives of the energy function $E$ w.r.t. parameters $t(x)$ and $J^c(x)$ for the case of SAD-norm are

$$\frac{\partial E}{\partial t(x)} = 2\sum_c \left(I^c(x) - t(x)J^c(x) - (1-t(x))A^c(x)\right)\left(A^c(x) - J^c(x)\right)$$

$$+ 2\lambda_1 \sum_{y \in N_x} \text{sgn}(t(x) - t(y)) + 2\lambda_3 \sum_c R^c(x) \quad (22)$$

$$\frac{\partial E}{\partial J^c(x)} = 2\left(I^c(x) - t(x)J^c(x) - (1-t(x))A^c(x)\right)(-t(x))$$

$$+ 2\lambda_2 \sum_{y \in N_x} \text{sgn}(J^c(x) - J^c(y)) \quad (23)$$

The closed solution of $w_i^c$ can be derived by the least square method,

$$w_i^c = \frac{\sum_x \left(I^c(x) - t(x)J^c(x)\right)(1 - t(x))g_i(x)}{\sum_x (1-t(x))^2 g_i^2(x) + \lambda_4(1 - \delta(i))} \quad (24)$$

where $\delta(i)$ is the Dirac delta function, it is 1 when $i = 0$ and is 0 otherwise.

The image dehazing result based on the airlight field estimation is shown in Figure 7. Compared with Figure 2 and 6, this image is uniform and vivid in color, due to the luminance compensation by the airlight field.

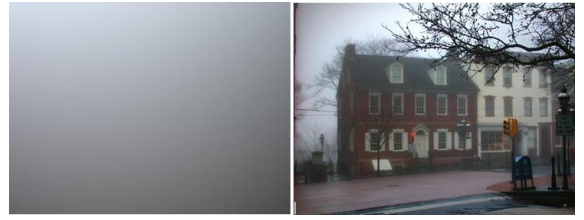

Figure 7: ALF estimation. Left: estimated airlight field map. Right: recovery image.

### 4.3. Image restoration

The steps of image dehazing algorithm based on airlight field estimation are as follows:

1) Initialization. Set $t = 0$, $J = I$, $w_i^c = 0$, for $i = 0 \cdots M - 1$. The parameters are $\lambda_1 = 0.1$, $\lambda_2 = 0.0001$, $\lambda_3 = 1$, $\lambda_4 = 0.1$. The step size is $\Delta t = 0.1$.
2) Find $A^c(x)$ as in Equation (20).
3) Update $t(x)$. Find $t_{k+1}(x) = t_k(x) - \Delta t \frac{\partial E}{\partial t_k(x)}$ as in Equation (22).
4) Update $J^c(x)$. Find $J_{k+1}^c(x) = J_k^c(x) - \Delta t \frac{\partial E}{\partial J_k^c(x)}$ as in Equation (23).
5) Update $w_i^c$ as in Equation (24).
6) Determine the termination of iteration. If either convergence has been reached or exceeds a predetermined maximum number of iteration, stop the iteration; otherwise, go to Step 2.

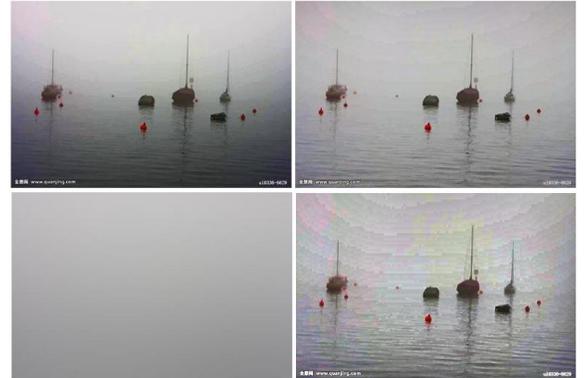

Figure 8: Our recovery image. TopLeft: recovery image by CBR. TopRight: recovery image by ALF. DownLeft: estimated airlight field. DownRight: recovery image by Equation (4).

Our recovery of the true image $J$ is combined with the estimation of the airlight field and transmission rate in the framework of energy minimization. On the contrary, many previous algorithm restore the true image by solving Equation (4) [14,16,17,20]. We found that they are prone to noise and quantification error in sky regions, when the transmission is close to zero. In fact, most outdoor photos contain sky regions that cannot be ignored. As shown in Figure 2, their restored images contain terrible halo, staircase effects and color distortion.

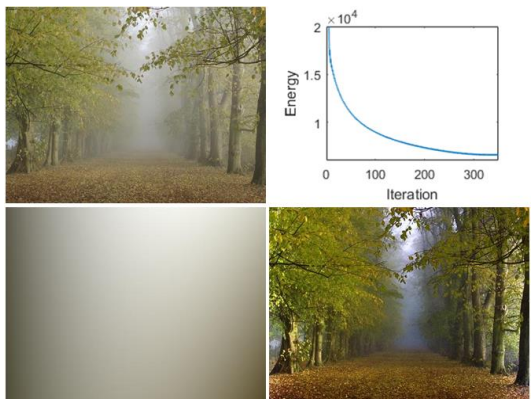

Figure 9: Airlight field dehazing and its energy curve. TopLeft: input hazy image. TopRight: energy curve. DownLeft: estimated airlight field. DownRight: recovery image.

As shown in Figure 8, our results have almost no such artifacts, both for constant background radiation (CBR) and airlight field (ALF) estimation, although the brightness of CBR is not uniform enough,. In order to verify our analysis about the reason of the halo, we use Equation (4) to get the final output image in our algorithm with airlight field estimation. The result is also shown in Figure 8. It contains severe haloes, which validates our analysis.

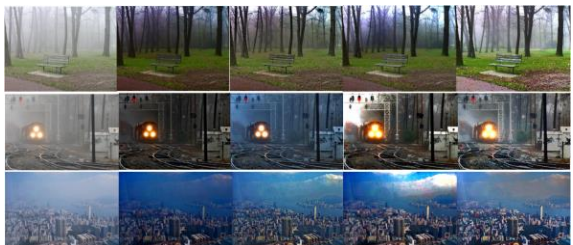

Figure 10: Comparison of results for non-constant airlight cases. From left to right: input haze image, results of He, Meng, Sulami, and ALF.

5. Experimental results

In our experiments, we use the 2-order Legendre polynomial as basis functions to estimate the airlight field. Set $t = 0$, $J = I$, $w_i^c = 0$, for $i = 0 \cdots M - 1$. The parameters are $\lambda_1 = 0.1$, $\lambda_2 = 0.0001$, $\lambda_3 = 1$, $\lambda_4 = 0.1$. The step size is $\Delta t = 0.1$. We use the gradient descent method as our solver, which iterates 200 times to get the satisfactory results. The energy curve is drawn in Figure 9, together with the estimated airlight field and the restored haze free image.

It takes about 1 minutes for Matlab code to process a 640x360 image on a PC with a 3.0 GHz Intel Pentium 4 Processor.

Figure 10 shows the case where the global atmospheric light is not constant. It intends to compare the proposed airlight field dehazing approach with He's, Meng's and Sulami's on several images [14,16,20]. In their output images, there is always one side brighter than the other side. Some images are dim on the whole, due to a false estimate of *A*. On the contrary, our results have uniform brightness.

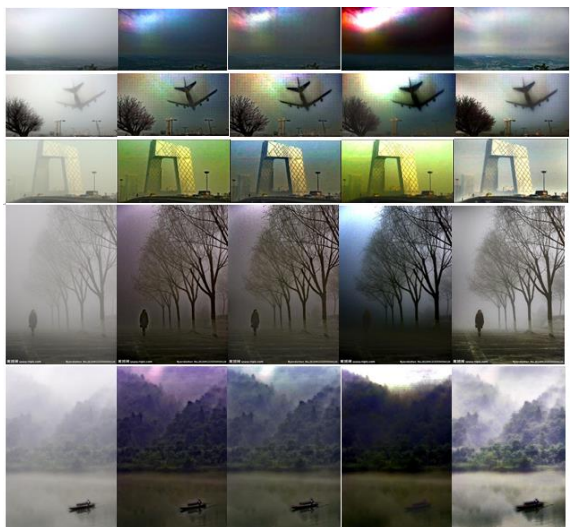

Figure 11: Comparison of results for images with large sky region. From left to right: input haze image, results of He, Meng, Sulami, and ALF.

In Figure 11, we further compare various methods on images which have large sky region. It is a difficult case for other algorithms to remove haze. As shown in the figure, all their algorithms suffer from terrible haloes, brightness imbalance and color distortion, and their results are completely unacceptable. Our technique yet generates perfect good haze free images.

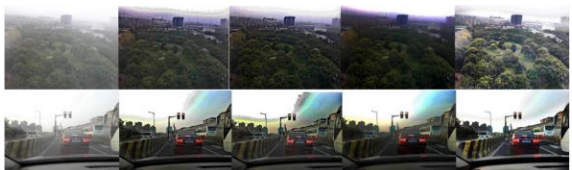

Figure 12: Comparison of staircase effects. From left to right: input haze image, results of He, Meng, Sulami, and ALF.

Figure 12 doesn't seem to be difficult to deal with. However, all other algorithms suffer from different degree of staircase effects with forged color in sky region, which we believe to be induced by noise and quantization error at small transmission points. Compared with them, our algorithm obtains much comfortable haze free images.

Our approach also works for colorless images very well. Figure 13 shows an example for comparison with other algorithms. The results prove that it is difficult for other algorithms to handle colorless images with large sky region. All of them suffer from obvious noise and forged color. As said by He, their approach can handle colorless images if there are enough shadows [14]. However, if this condition is not satisfied, it does not perform well.

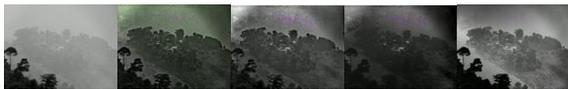

Figure 13: Comparison of results for colorless images. From left to right: input haze image, results of He, Meng, Sulami, and ALF.

There are still drawbacks in our algorithm. Besides the large computation time, it is the distortion of the estimated airlight field. When the light and shade contrast of the image is too strong, it will artificially cause some areas in the recovery image to be too bright, e.g., as shown in Figure 14. This is the result of underfitting for the airlight field function. Because the number of basis functions is too small, the abrupt change of airlight field is difficult to be traced accurately. Figure 14 shows the case of fitting the airlight field by 2, 3 and 4 order polynomial functions. An obvious trend is that when using high-order polynomials as basis functions, the distortion of the recovery image is small. Therfore, we should use the polynomial function as high as possible. However, this will further increase the amount of calculation.

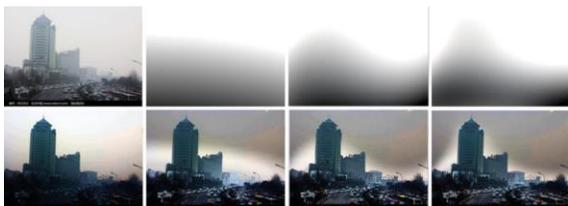

Figure 14: Distortion of estimated airlight field. Top: input hazy image, estimated airlight fields with 2,3,4 order polynomial functions. Down: recovery image with CBR, and with ALF by 2,3,4 order polynomial functions.

## 6. Discussion

Conventional dehazing approaches cannot handle hazy images which contain large sky region and may fail when the constant-airlight assumption is broken. It induces brightness imbalance, color distortion, halo and staircase effects in recovery images. To address this problem, we propose the haze removal algorithm based on the airlight field (ALF) estimation.

While the proposed approach achieves remarkable results, the main defect is the large time consuming. One solution that can be considered is to give an initial value that is fully close to the optimal solution, followed by the conjugate descent or quasi-Newton method instead of the gradient descent method to accelerate the convergence speed.

Another drawback of the proposed method is the distortion of the estimated airlight field caused by underfitting. This phenomenon frequently occurs when the light and shade contrast of the image is too strong. We expect that if the higher order polynomial is used as the basis function, this symptom will be alleviated. This will dramatically increase the amount of computation and storage. Therefore, we need in the future to find a set of more suitable basis function and reasonable order to get the trade-off between fitting accuracy and computation amount.

We have also considered using the total variation(TV) regularizaiton for $t$ and $J$ in the energy mimization formula instead of the SAD-norm regularization. There are a variety of effective methods for speeding up the convergence of the variational optimization, including [21-23]. TV regularization has proven to be particularly useful to solve a number of ill-posed inverse imaging problems, such as image denoising and deblurring. It is also used in haze removal by Zhou in [18]. We leave this issue for future research.

## 7. Conclusion

In this paper, we propose a new dehazing approach by modeling the global atmospheric light as a airlight field function. In the proposed model, a given set of smooth orthogonal basis functions are used to estimate the slowly varying airlight field with a linear combination. We also introduce a haze-level prior to help estimate the transmission rate. The recovery image is generated in a new energy minimization formula without the brightness imbalance, color distortion, halo and staircase artifacts, which are inherent in the previous algorithms when dealing with large sky images. The experimental results indicate that our approach achieves superior satisfactory haze removal effects, when compared with state of the art approaches.